\documentclass[cha, reprint]{revtex4-1}

\usepackage{times}
\usepackage{graphicx}

\begin{document}

\title{Analysis of Multi-Scale Fractal Dimension to Classify Human Motion}

\author{N\'ubia Rosa da Silva}
\email{nubiars@icmc.usp.br}
\affiliation{Institute of Mathematical Sciences and Computing - ICMC\\
University of Sao Paulo - USP}
\author{Odemir Martinez Bruno}
\email{bruno@ifsc.usp.br}
\affiliation{Physics Institute of Sao Carlos - IFSC\\
University of Sao Paulo - USP}

\begin{abstract}
   In recent years there has been considerable interest in human action recognition. Several approaches have been developed in order to enhance the automatic video analysis. Although some developments have been achieved by the computer vision community, the properly classification of human motion is still a hard and challenging task. The objective of this study is to investigate the use of 3D multi-scale fractal dimension to recognize motion patterns in videos. In order to develop a robust strategy for human motion classification, we proposed a method where the Fourier transform is used to calculate the derivative in which all data points are deemed.  Our results shown that different accuracy rates can be found for different databases. We believe that in specific applications our results are the first step to develop an automatic monitoring system, which can be applied in security systems, traffic monitoring, biology, physical therapy, cardiovascular disease among many others.
\end{abstract}

\maketitle

\section{Introduction}

In recent years, there has been a growth of research activity aimed to develop human motion classifiers in order to enhance the automatic video analysis. Several approaches for tracking movement have been proposed in the literature.  Basically, they differ in the type of object representation, varying size, position and shape of moving objects, in applying the type of motion and appearance model.

All these perspectives are designated according to the context and end-use monitoring will be conducted. Regarding the context in which the movement can be recognized, different and interesting applications had been considered. For instance, human-computer interface, gesture recognition, video indexing and browsing, analysis of sports events and video surveillance. In these situations, to recognize events of particular interest as well as their complexity and to make inferences about their evolution play a crucial rule in image processing research. All these tasks can be done by considering the knowledge that can be obtained from the motion patterns.

Although developments have been achieved by the computer vision community, the properly classification of human motion is still a hard and challenging task. This because (i) it is not possible to control the acquisition of images sequence; (ii) the images can suffer from poor illumination, blur, occlusion or of several other possibilities. Moreover, in real world, the situations differs a lot from the controlled conditions tested at the laboratory.

An appropriate method is required to assess the motion in different videos. In this study we focus on classification of the following single human motions: ``walk'', ``skip'', ``kick'', ``playing basketball'', ``run'', ``jack'', ``jump'', side and  ``wave'' under unconstrained indoor environments. The purpose of this paper is to investigate the use of 3D multi-scale fractal dimension to recognize motion patterns in videos. In order to study and develop a more robust strategy for human motion classification, we are using the Fourier Transform to calculate the derivative in which all data points are deemed instead of numerical methods.

Each motion class is characterized by a signature obtained by multi-scale fractal dimension-based approach. Different motions will provide distinct signatures, therefore we can discern dissimilar motions. The multi-scale fractal dimension  was performed by using the Bouligand-Minkowski method, a robust, accurate and consistent way to estimate the fractal dimension according to literature \cite{Backes08fractaland,tricot1995,luciano2000_book}. The motion signature is a curve of multi-scale fractal dimension that represents the changes in shape complexity frame by frame for different scales observed.

The rest of paper is organized as follows. Extraction signatures by multi-scale fractal dimension is explained in Section \ref{sec:methods}. Experimental results and discussions are presented in Section \ref{sec:results} and conclusions in Section \ref{sec:conclusions}.

\section{Signatures by Multi-Scale Fractal Dimension}\label{sec:methods}

Fractal analysis has been widely applied to describe different problems in pattern recognition, image processing and many others domains. Established by Benoit Mandelbrot the fractal geometry is useful in problems that require complexity analysis of structures across different scales \cite{mandelbrot1983fractal}. It is necessary to point out that the metric properties of fractal objects are a function of the scale used to perform the measurement. Then, we can describe an object with fractional values depicting the level of complexity and spatial distribution  in the image \cite{luciano2000_book,falvo2005,Backes2010c,Lopes_Betrouni2009}.

\subsection{Fractal Dimension}

The fractal dimension indicates how much space is occupied by the object, representing the degree of complexity that the figure has. 
For uniform and compact objects the fractal dimension coincides with the topological dimension, however, for fractal objects it is a fractional value. To estimate the fractal dimension of an object several methods can be used, including box-counting, mass-radius, dividers and Bouligand-Minkowski approach. For this study we take the Bouligand-Minkowski method into account due the precision and adaptation to the multi-scale approach \cite{tricot1995,luciano2002,Backes2010d}. Usually Minkowski is used for many applications in shape and texture analysis. While in shape, the Minkowski approach considers two-dimensional space, for texture we usually handle with three-dimensional space, due the third coordinate corresponds to the intensity of gray level at each point. In our case, we also have a three-dimensional space where the third coordinate is the time, i.e., the sequence of images in which actions occur during the video.

\begin{figure}[h]
\centering
\includegraphics[height=5.4cm]{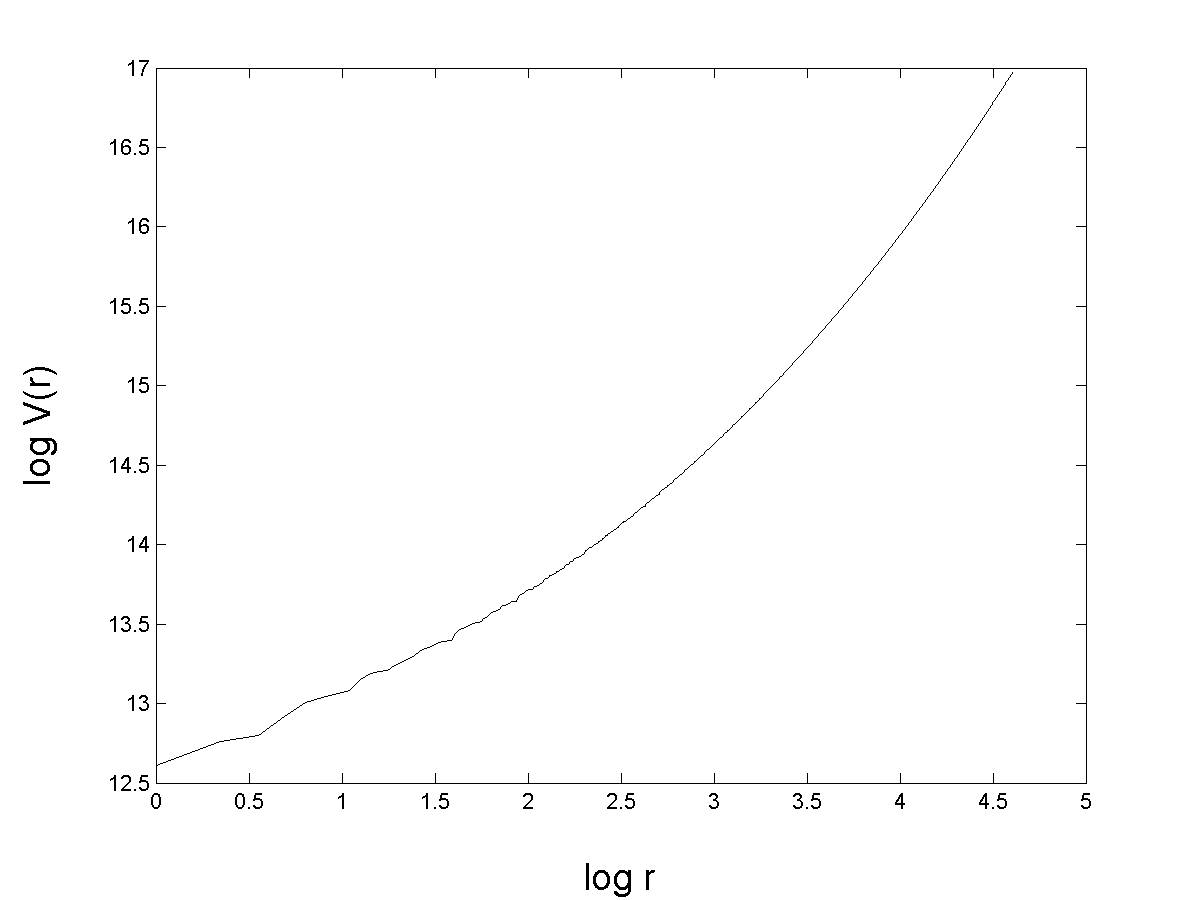} 
\caption{Log-log curve of a sequence of video images generated by Bouligand-Minkowski.}
\label{fig:loglog}
\end{figure}

By using Bouligand-Minkowski we analyze the relationship between the object and the space occupied by it in space. The fractal dimension is obtained by calculating the volume of the dilated object. The dilatation can be performed by considering a sphere of radius $r$ (Figure \ref{fig:dilata3D} a)), which is centred at each point of the original object and all other points inside the sphere are joined to the object. In order to generate a signature for the shape, we analyzed the object volume as a function of $r$. The algorithm used to perform this task consists in use the exact distance transform (EDT) \cite{odemirLuciano2004,Saito1994,Fabbri2008acm,JulioFabbri}, which is the distance of all points of the image to the closest point of the object. After that, we computed the fractal dimension by analyzing the log-log curve of the volume of influence \emph{versus} $r$ (Figure \ref{fig:loglog}). The fractal dimension is defined as:

\begin{equation}\label{eq:fd}
    FD(r) = N - \frac{\log A(r)}{\log r},
\end{equation}

$\ \\$where $A(r)$ is the influence volume of the object with radius $r$ and $N$ is the number of dimensions.

Over the video frames it is possible find different visions of the same object. In a image of the video sequence the motion object can be close to the camera while in another image/frame can be far from the camera. Because of the range of scales of the object during the image sequence and for  have just a value for the whole video, we use an extension of fractal dimension that is Multi-Scale Fractal Dimension.

\subsection{Volumetric Multi-Scale Fractal Dimension}

The log-log curve produced by the Bouligand-Minkowski method presents a wealth of detail that can not be represented by  only one value provided by the fractal dimension. For this reason the multi-scale fractal dimension uses the derivative to explore the limit of infinitesimal linear interpolation. Thus it is possible to obtain the relationship between variations in the complexity of the object at different scales \cite{Torres2004,falvo2005}. Multi-scale fractal dimension curve is defined through the derivative  of $\log V(r)\times \log r$ curve. For computing the influence volume we use the Boulingand-Minkowski method for three dimensions and we use the derivative property of Fourier transform to calculate the derivatives.	

\begin{figure}[h]
\centering
\includegraphics[height=5cm]{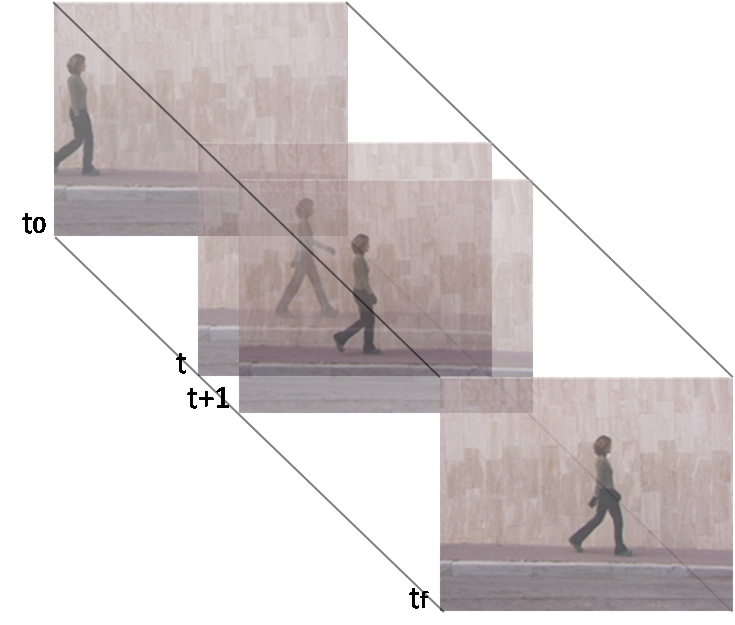}
\caption{Temporal variation in the sequence of video images.}
\label{fig:seq_tempo}
\end{figure}

In the proposed method, each image $ I \in R^2 $ of the sequence of images (video) is considered as a surface $S \in R^3 $. Each pixel of the image is converted to a point $ p = (x, y, z) $, $ p \in S $, where $x$ and $y$ are the coordinates of the object in the image $n \times n$, and $ z $ is the frame  in the sequence of $t$ images, as shown in Figure \ref{fig:seq_tempo}.

\begin{figure}[h]
\centering 
\includegraphics[height=3.5cm]{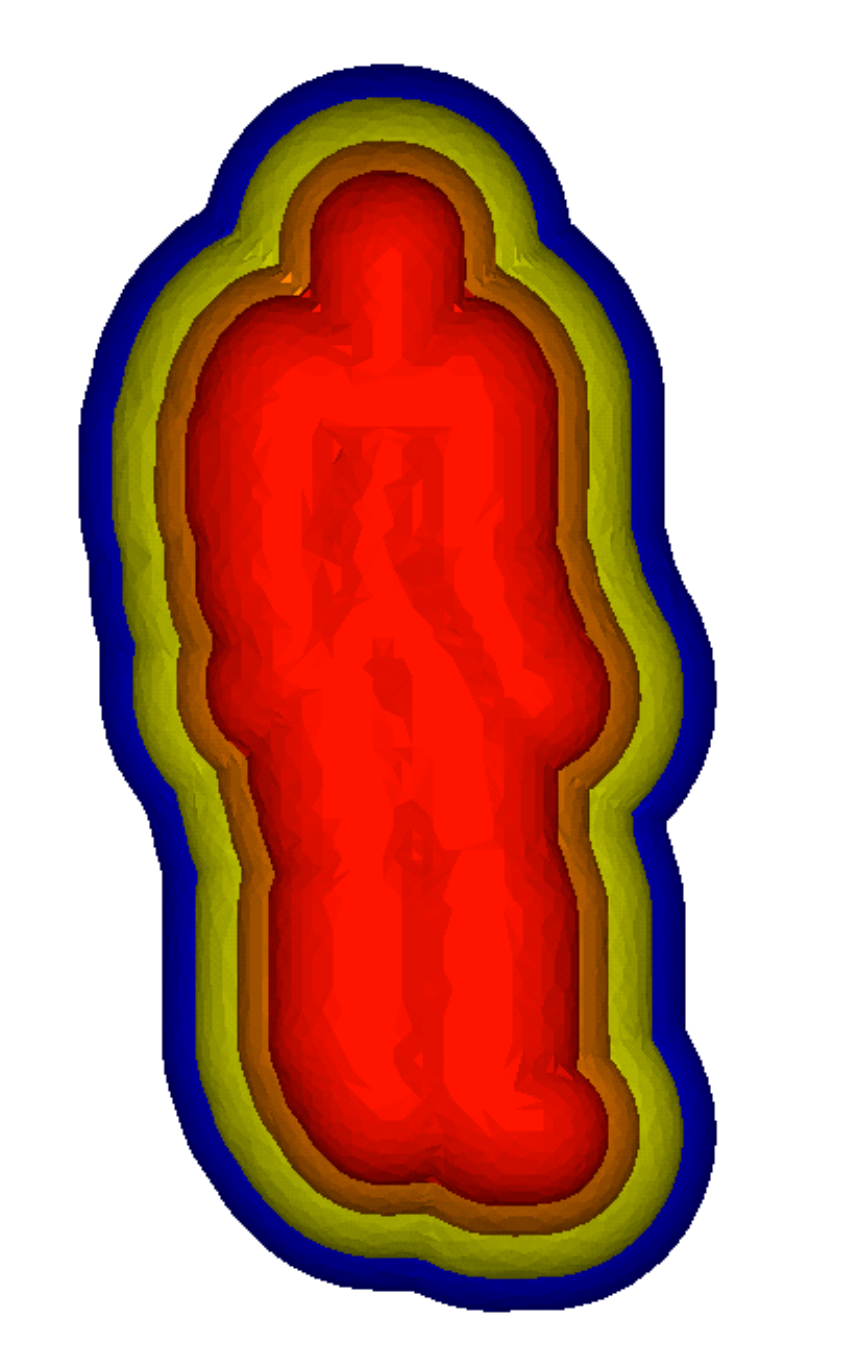}
\includegraphics[height=3.5cm]{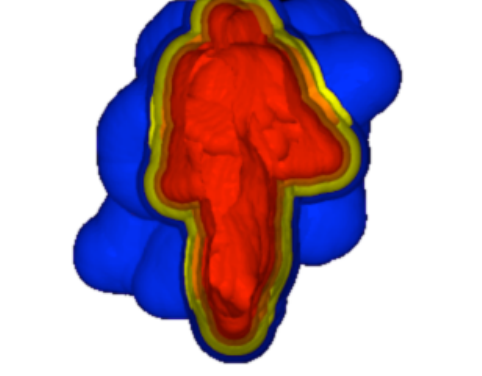} \\
\noindent a \hspace{2.5cm} b
\caption{Dilation of shape. a) 2D and b) in time. Red indicates radius $r = \sqrt{50}$,  orange indicates $r = \sqrt{100}$, yellow with $r = \sqrt{200}$ and  blue $r = \sqrt{300}$.}
\label{fig:dilata3D}
\end{figure}

The volume of the dilated shape in time $V(r)$ (Figure \ref{fig:dilata3D} b)) using a sphere of radius $r$ can be written as:

\begin{equation}\label{eq:dilatacao}
    V(r) = \sum_{i = 0}^{N}  \Theta(r, r_i),
\end{equation}

$\ \\$where $r_i$ is the minimum distance from a point $i$ to any other point $p$ belonging to the object, $N$ is $n \times n \times t$ and $\Theta(r, r_i)$ is the Heaviside function \cite{Butkov1968_book} that returns 1 if $r \geq r_i$ and 0, otherwise.

We compute the Bouligand-Minkowski fractal dimension $D$ as:

\begin{equation}\label{eq:dimfractal3}
    D = 3 - \lim_{r \rightarrow 0} \frac{\log V(r)}{\log r}
\end{equation}

As it has been considered three-dimensional space, $D$ is within $[0,3]$ and 3 is number of dimensions. According to the radius $r$, the volume of a sphere produced by a point $p \in S$ affects the volume of other spheres, disturbing the way the volume of influence increases. This makes the volume of influence $V(r)$ very sensitive to structural changes \cite{dalcimarBackes2009}.

Thereafter, the Multi-scale fractal dimension were taken using:

\begin{equation}\label{eq:mfd}
    MFD = 3 - \frac{d \log V(r)}{d \log(r)}
\end{equation}

\begin{figure}[h]
\centering
\includegraphics[height=5.4cm]{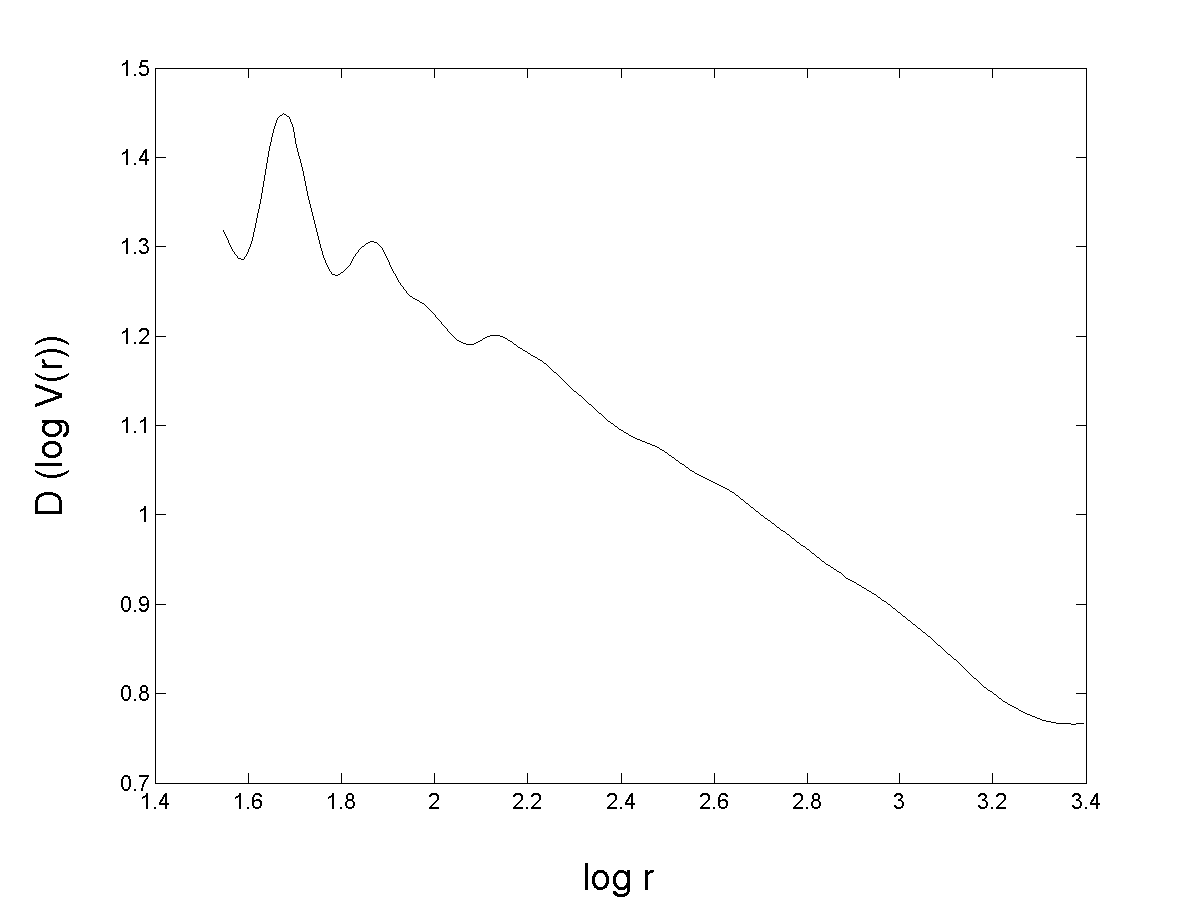}
\caption{Multi-scale Fractal Dimension of a sequence of video images.}
\label{fig:dfm}
\end{figure}

The result is a curve with the fractal dimension calculated at each spatial scale represented by the radius $r$ (Figure \ref{fig:dfm}). The curves of the multi-scale fractal dimension were considered as signatures for the study of the motion patterns. 

To calculate the derivative we use a property of Fourier transform, which allow us to obtain the derivative of any function from the analysis of its spectrum of frequencies \cite{DigImageProc2ndEdGonzalez}. We also use a convolution of the original signal with a Gaussian kernel in order to smooth the derivative. 

\begin{figure}[h]
\centering
\includegraphics[height=5.4cm]{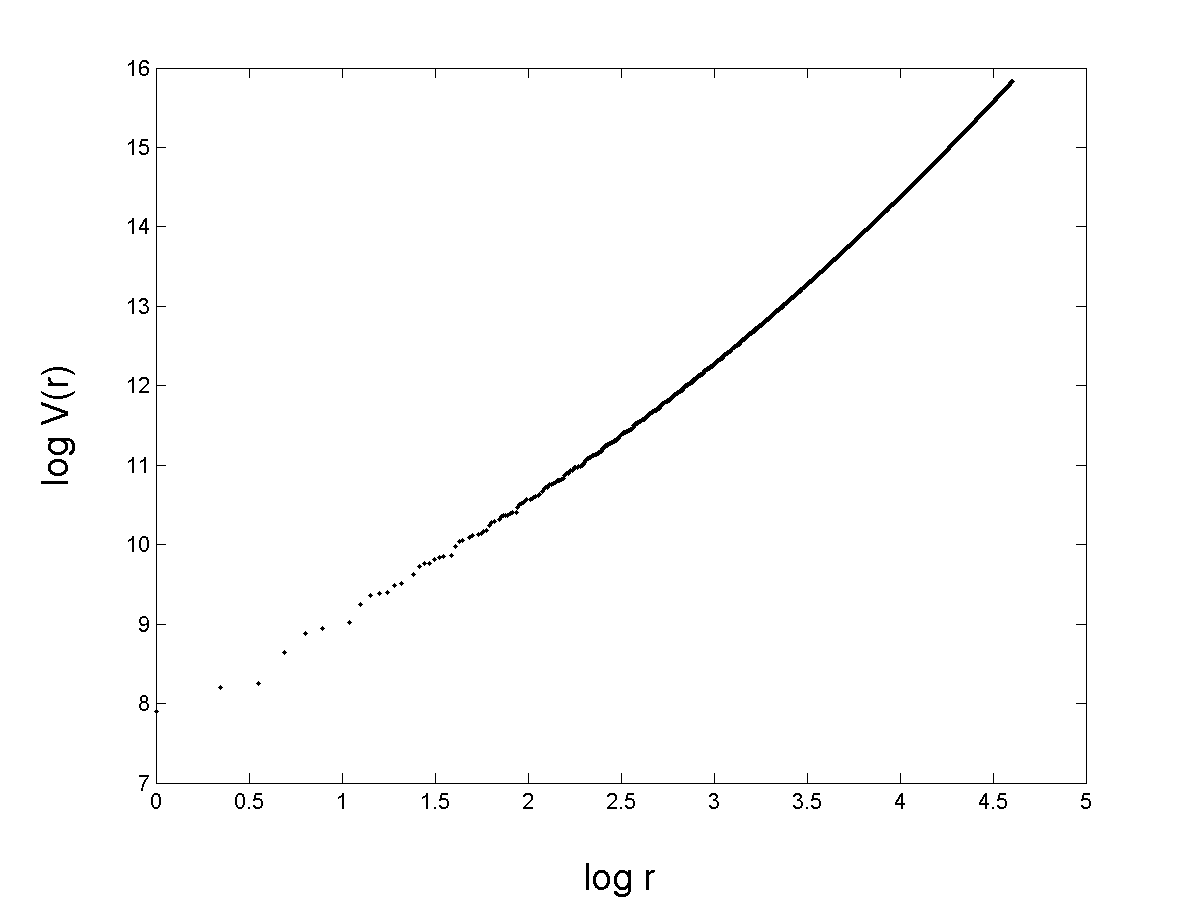}
\caption{Very sparse points are ignored.}
\label{fig:log-logSparso}
\end{figure}

\begin{figure}[h]
\centering
\includegraphics[height=5.4cm]{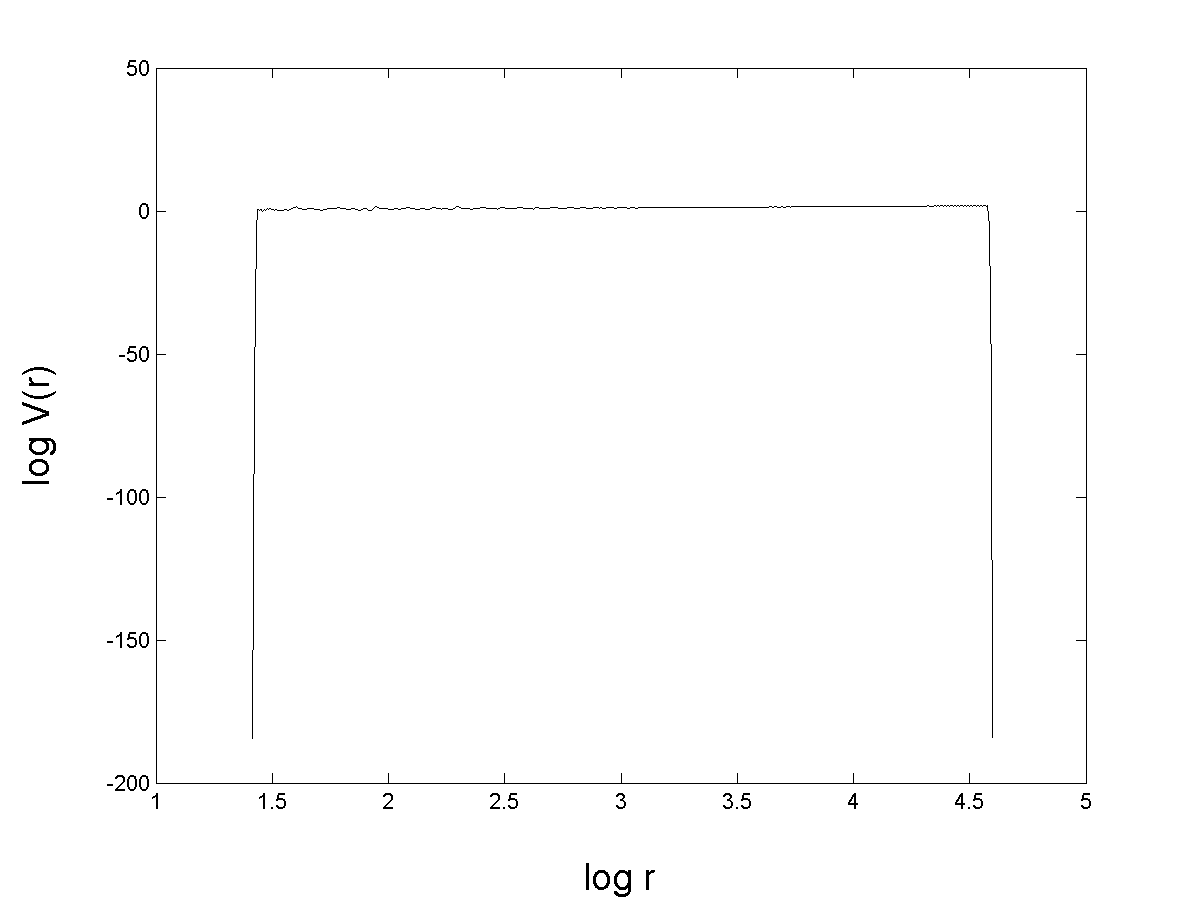}
\caption{Fourier transform does not converge uniformly in discontinuities causing the so-called Gibbs phenomenon.}
\label{fig:gibbs}
\end{figure}

\begin{figure}[h]
\centering
\includegraphics[height=5.4cm]{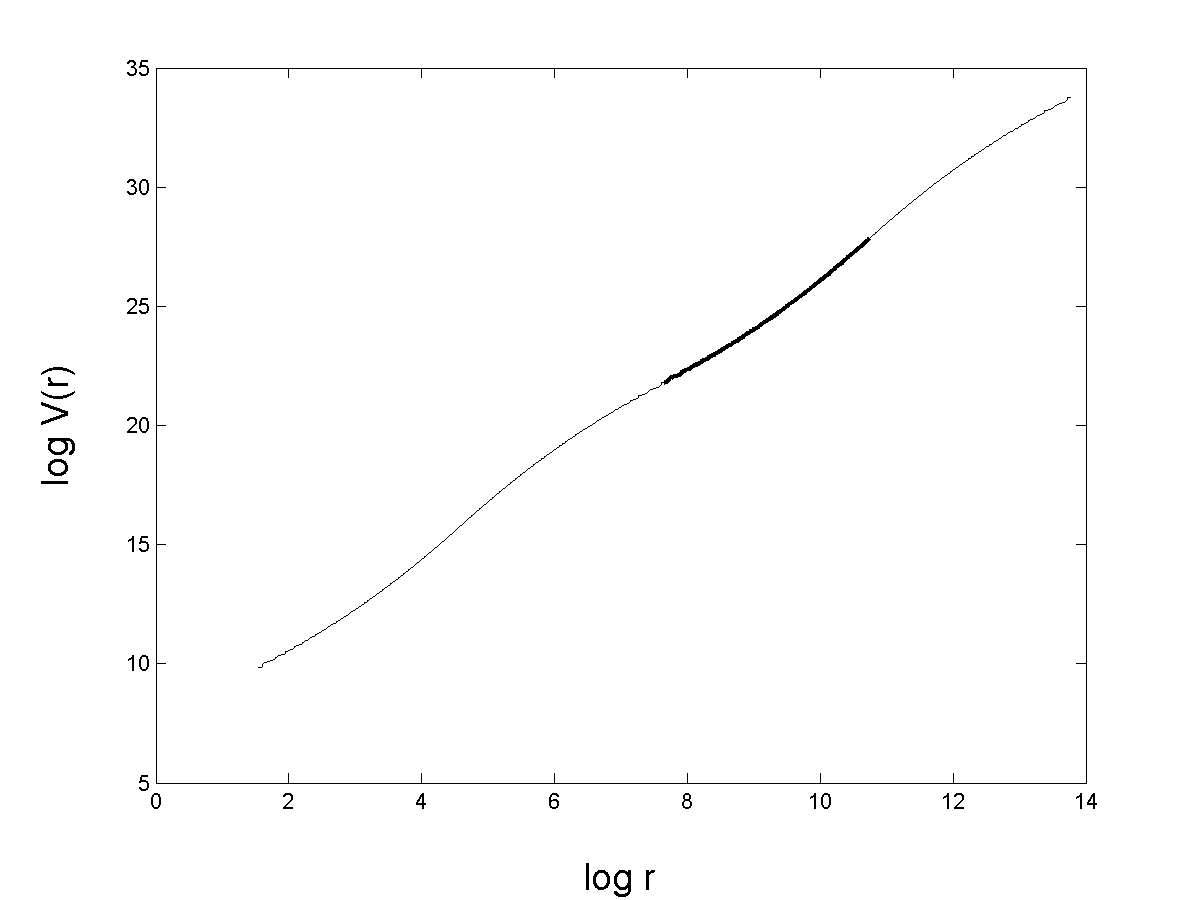}
\caption{Curve replication  before and after the original curve to solve the Gibbs phenomenon.}
\label{fig:replica}
\end{figure}

Two questions have to be observed when calculating the derivative. The first one is the spacing between points of the signal, because the log-log curve has a very low sampling at the beginning, as showed in Figure \ref{fig:log-logSparso}. 
The sparse points were ignored and the remaining points were interpolated by filling spaces between each two points with their average. The second question is that the Fourier transform does not converge uniformly in discontinuities
causing the so-called Gibbs phenomenon at the ends of the signal (Figure \ref{fig:gibbs}). 
To solve this problem, the curve was replicated before and after the original curve (Figure \ref{fig:replica}).

\section{Results and Discussions}\label{sec:results}

The analysis of motion is still a challenging problem in the field of image processing and pattern recognition. Over the years, several approaches using different constructs have been proposed for action recognition, including: machine learning \cite{Minhas2010,Cao2009}, optical flow \cite{Horn1981,Denman2005,Roberts2009}, appearance model \cite{eraldoICIAR2009,eraldoCVPR2008,zhaoTPAMI2008}. They differ especially in the kind of object representation, image features, and in applying the type of motion. 

In this paper we have studied a new strategy to characterize motion in a image sequence, as well as to recognize the motion patterns in order to classify the movement. First, we use only the fractal dimension in the three-dimensional space to characterize the motion. However, it was unable to capture structural differences in the shape of the moving object. Another problem is that the method was not scale-invariant. 
To overcome this problem, we investigate a shape in time over multiple scales. For each scale we have a fractal dimension and the set of consecutive fractal dimensions is called signature of the video sequences. The main goal of this paper is to use these signatures in order to discriminate different types of movement.

\begin{figure}[h]
\centering
\includegraphics[height=5.4cm]{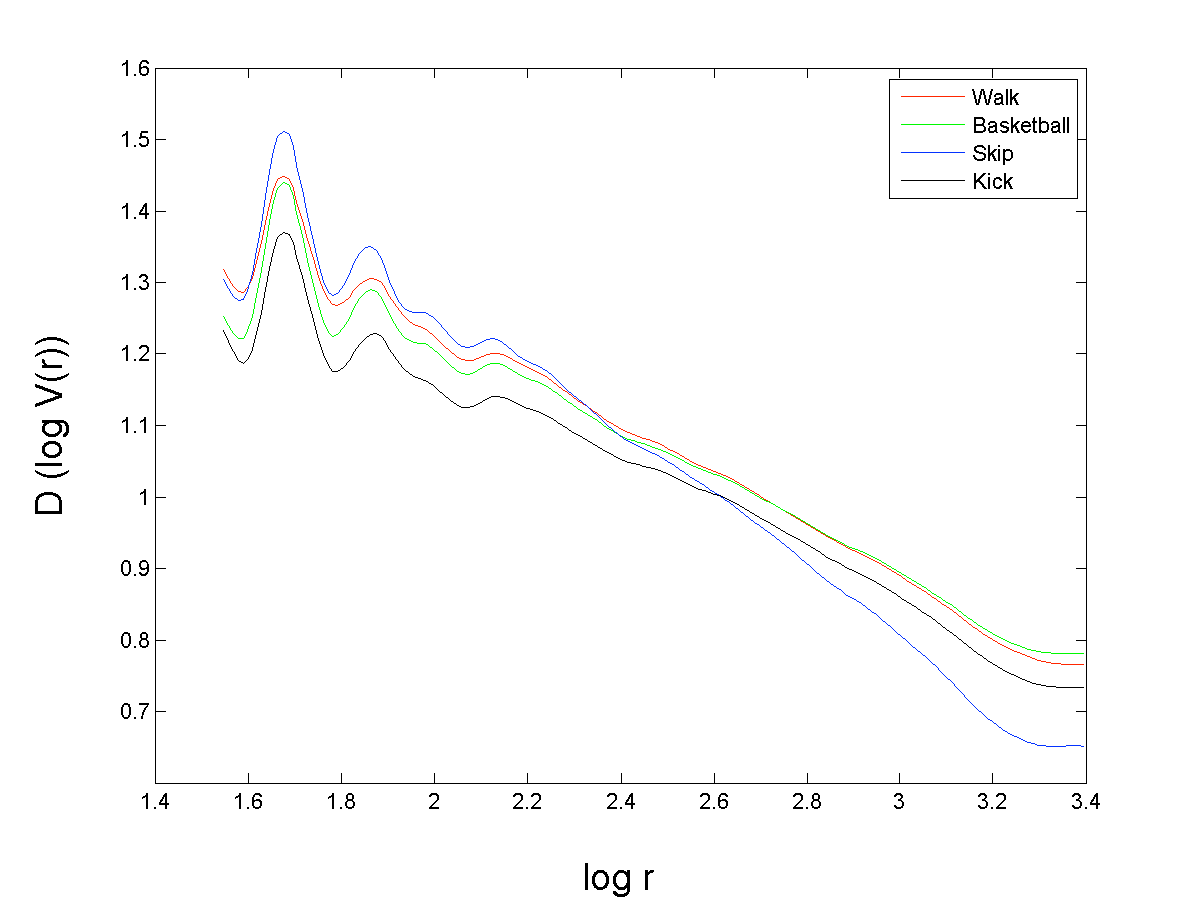}
\caption{Motion signature by Multi-scale Fractal Dimension of four different movements.}
\label{fig:dfmmotion}
\end{figure}

To confirm the hypothesis that motion signatures can be distinct for different movements, we calculate the multi-scale fractal dimension to all actions. We compared the similarity of the same type of movement  to ensure that signatures could be obtained similar to movements of the same class and different signatures for movements belonging to different classes. Figure \ref{fig:dfmmotion} shows an example of four different motion signatures.

Basically, there are two parameters in our approach, $\sigma$ and $r_{max}$. The first one is related to the standard deviation of the Gaussian kernel and it quantifies the level of smoothness of the signatures. The last one is related to the maximum scale that is used to analyze an image sequence. We investigate the effects of this two parameters and we conclude that for very high values of $\sigma$, we have a strongly smooth curves which implies loss of important details of the signature. We tested $\sigma$ varying from 1 until 6 and the maximum radius equals 160. Regarding the parameter $r_{max}$, to find out the optimal value becomes difficulty, because it is not known a priori at what scale the motion must be analyzed.

Our strategy was demonstrated performing tests on two publicly available dataset, CMU Graphics Lab Motion Capture Database (available at \emph{http://mocap.cs.cmu.edu/}) and Weizmann Human Action Dataset \cite{ActionsAsSpaceTimeShapes_pami07,ActionsAsSpaceTimeShapes_iccv05} (\emph{http://www.wisdom.weizmann.ac.il/~vision/SpaceTimeActions.html}). 

A signature was then generated for each video and we classify the motions according to their signatures. The result corresponds to a feature vector based on the signature of the multi-scale fractal dimension and each point of the signature will be an attribute.  Support vector machine with 10-fold cross validation scheme was chosen to classify the examples. SVM uses the Statistical Learning Theory, first assuming the data domain in which learning is occurring are generated independently and identically distributed according to a probability distribution relationship between the examples and classes \cite{Chen2005}. Thus, to new data from the same domain, SVM obtains good results. The result of  10-fold cross validation is the average values for 10 runs of the same experiment through random selection of  actions in the database. 

We accomplished two sets of experiments. The first one on Mocap database. And the second one on Weizmann Database. All experiments were performed using Weka \cite{weka} with default values parameters.

\subsection{Mocap Database} 

The Motion Capture Database contains 2605 trials in six motion categories and 23 subcategories. We choose 66 video sequences showing eight different subjects which perform four distinct actions at varying speeds. The actions are: ``walk'', ``skip'', ``kick'' and ``playing basketball''. The cameras are placed around a rectangular area, of approximately 3 m x 8 m, in the center of the room. Only motions that take place in this rectangle can be captured. The Mocap database contains videos with 53 a 1020 images of size 240 $\times$ 320.

To find the best values for $r_{max}$ we performed tests varying the radius from 10 to 160 and $\sigma$ equals 1. 
The number of  correctly classified instances is almost constant by varying the size of the radius, this because of the size of each video image. Then, we can use the lowest.

To evaluate the better $\sigma$ value  we varied it from 1 until 6. 
According to which the $\sigma$ value increases the number of correctly classified instances decreases, as the high smooth of motion signature. 

From the performed tests we conclude the better values are $r_{max}$ equals 10 and $\sigma$ equals 1. It was obtained 90.91 \% (standard deviation equals 0.34) of accuracy with 60 of 66 correctly classified instances.

\subsection{Weizmann Database}

The Weizmann human action dataset has 93 video sequences showing nine different subjects, each performing 10 natural actions such as ``run'', ``walk'', ``skip'', ``jumping-jack'' (or shortly ``jack''), ``jump-forward-on-two-legs'' (or ``jump''), ``jump-in-place-on-two-legs'' (or ``pjump''), ``gallopsideways'' (or ``side''), ``wave-two-hands'' and  ``wave-one- hand'' (or ``wave''), or ``bend''. The Weizmann database contains videos with 28 a 146 images of size 144 $\times$ 180. The size of video are not correlated to the motion.
 
We tested different values for $r_{max}$ and $\sigma$. 
There is little variation in the number of correct classifications by varying the size of the radius, but we note that with a $r_{max}$ equal to 110, has the highest rating. 
With $r_{max}$ equals 110, we vary the value of $\sigma$ from 1 to 6. 
All combinations of radius and sigma have been tested, but only the most relevant values are shown here. Again, with increase of $\sigma$, the correctly classified decreases, the better result is obtained with $\sigma$ equals 2. Therefore, with radius equals 110 and $\sigma$ equals 2, the result of the classification of  Weizmann database is 79.57 \% (standard deviation equals 0.28) of accuracy with 74 of 93 correctly classified instances.

\section{Conclusions} \label{sec:conclusions}

This paper presented a study of motion classification based on a frame-by-frame analysis of the complexity of a shape in a video. Our main goal was apply the so-called multi-scale fractal dimension in order to classify videos according their content. We developed a strategy to classify human motion using multi-scale fractal dimension that consists to represent the movement contained in a video by a signature and support vector machine to classify. We have applied the method we describe in two real databases. The first one with 66 videos and four different types of motion and the second one with 93 videos and ten types of movements, where we have obtained different results with 90.91 \% and 79.57 \% of accuracy, respectively. The first database has only four motion classes quite different, however the second one has ten classes with some similar as the case of ``run'', ``side'' an ``skip''. In fact, it will be interesting to perform new experiments in a larger data set with the intention of finding a general strategy that can be potentially applied in a wide variety of vision problems that involve various complex structures of motion.

\section*{Acknowledgements}
N.R.S. acknowledges support from FAPESP (Grant \#2011/21467-9).\\ 
O.M.B. acknowledges support from CNPq (Grant \#308449/2010-0 and \#473893/2010-0) and FAPESP (Grant \# 2011/01523-1).


%

\end{document}